\documentclass[conference,10pt]{IEEEtran}
\usepackage{amsmath}
\usepackage{amssymb}
\usepackage{graphicx}
\usepackage{color}
\usepackage{cite}
\usepackage{bm}
\usepackage{epsfig,psfrag}
\usepackage{tabularx}
\usepackage{acronym}
\usepackage[yyyymmdd,hhmmss]{datetime}
\usepackage{bbm}
\usepackage{mathrsfs} 
\usepackage{bardiag}
\usepackage{tabularx}
\usepackage{multirow}
\usepackage{colortbl,pgfplotstable}
\usepackage{notation}
\usepackage{subfig}
\usepackage{float}
\usepackage{epsfig,psfrag}
\usepackage{soul}
\usepackage[inline]{enumitem}

\allowdisplaybreaks
\newcommand{\rd}{\textcolor{red}}

\newcommand{\be}{\begin{equation}}
\newcommand{\ee}{\end{equation}}
\newcommand{\ist}{\hspace*{.3mm}}
\newcommand{\rmv}{\hspace*{-.3mm}}

\newcommand{\nn}{\nonumber}
\newcommand{\T}{\mathrm{T}}

\newcommand{\acr}[1]{\aclu{#1} (\acs{#1})}

\definecolor{myblue}{RGB}{0, 0, 255}

\acrodef{pmf}[PMF]{probability mass function}
\acrodef{pdf}[PDF]{probability density function}
\acrodef{mmse}[MMSE]{minimum mean-square error}
\acrodef{gnn}[GNN]{graph neural network}
\acrodef{mpnn}[MPNN]{message passing neural network}
\acrodef{da}[DA]{data association}
\acrodef{bp}[BP]{belief propagation}
\acrodef{mlp}[MLP]{multi-layer perceptron}
\acrodef{fg}[FG]{Factor Graph}
\acrodef{nebp}[NEBP]{neural enhanced belief propagation}
\acrodef{roi}[ROI]{region of interest}
\acrodef{bev}[BEV]{bird's eye view}
\acrodef{mot}[MOT]{multi-object tracking}
\acrodef{iou}[IoU]{Intersection over Union}
\acrodef{po}[PO]{potential object}
\acrodef{fp}[FP]{false positives}
\acrodef{ids}[IDS]{identity switches}
\acrodef{frag}[Frag]{fragments}
\acrodef{amota}[AMOTA]{average multi-object tracking accuracy}
\acrodef{amotp}[AMOTP]{average multi-object tracking precision}
\acrodef{nms}[NMS]{non-maximum suppression}

\begin{document}

\title{\vspace{0mm} Neural Enhanced Belief Propagation\\ for Data Association in Multiobject Tracking \vspace{-1mm}}


\author{\IEEEauthorblockN{Mingchao Liang and Florian Meyer}
\IEEEauthorblockA{Department of Electrical and Computer Engineering, University of California San Diego, La Jolla, CA}\\[-4.6mm]
Email: \{m3liang, flmeyer\}@ucsd.edu\vspace{0mm} }


\maketitle


\begin{abstract}
Situation-aware technologies enabled by \ac{mot} methods will create new services and applications in fields such as autonomous navigation and applied ocean sciences. Belief propagation \acused{bp}(\ac{bp}) is a state-of-the-art method for Bayesian \ac{mot} that relies on a statistical model and preprocessed sensor measurements.
In this paper, we establish a hybrid method for model-based and data-driven \ac{mot}. The proposed \ac{nebp} approach complements \ac{bp} by information learned from raw sensor data with the goal to improve data association and to reject false alarm measurements. We evaluate the performance of our \ac{nebp} approach for \ac{mot} on the \textit{nuScenes} autonomous driving dataset and demonstrate that it can outperform state-of-the-art reference methods.
\vspace{1mm}
\end{abstract}

\begin{IEEEkeywords}
Belief propagation, graph neural networks, multiobject tracking,\vspace{-1mm} factor graph.
\end{IEEEkeywords}

\section{Introduction}

Multi-object tracking \acused{mot}(\ac{mot}) \cite{BarWilTia:B11} is a key aspect in a variety of applications including autonomous navigation and applied ocean sciences. In particular, in autonomous navigation systems accurate MOT enables tasks such as motion forecasting \cite{LeeChoVerChoTorCha:17} and path planning \cite{ZenLuoSuoSadYanCasUrt:19}. The main challenge in \ac{mot} is data association uncertainty, i.e., the unknown association between measurements and objects. MOT is further complicated by the fact that the number of objects is unknown. 


Most \ac{mot} methods adopt a detect-then-track approach where an object detector \cite{ZhuTuz:18, LanVorCaeZhoYanBei:19, ZhuJiaZhoLiYu:19, YinZhoKra:21, RenHeGirSun:15, ShiWanLi:19, SimBulPorLop:19,BarLiuWinCon:J22,MeyGem:J21,BraGagSolMenFerLepNicWilBraWin:J22} is applied to the raw sensor data. The resulting object detections are then used as measurements for \ac{mot}. 
Many existing \ac{mot} methods follow a global nearest neighbor \cite{BarWilTia:B11} approach where a Hungarian \cite{Kuh:55} or a greedy matching algorithm is used to perform ``hard'' data association. These types of methods rely on heuristics for track initialization and termination. To improve the reliability of hard data association, often discriminative features are extracted and incorporated by the matching algorithm \cite{WenWanHelKit:20, YinZhoKra:21, WenWanManKit:20, ChiLiAmbBoh:21, ZaeDaiLinDanVan:22, RanMahGebMhaRamTri:21, WanZheLiuLiWan:20, ZhaWanWanZenLiu:21,LiaYanZenCheHuCasUrt:20}. Another line of work formulates and solves MOT in the Bayesian estimation framework \cite{BarWilTia:B11,Wil:J15,MeyBraWilHla:J17, MeyKroWilLauHlaBraWin:J18, GarWilGraSve:J18, SchBenRosKriGra:18,PanMorRad:21,ZhaMey:C21}. This type of methods  rely on statistical models for object birth, object motion, and sensor measurements \cite{BarWilTia:B11,Wil:J15,MeyBraWilHla:J17, MeyKroWilLauHlaBraWin:J18, GarWilGraSve:J18, SchBenRosKriGra:18,PanMorRad:21,ZhaMey:C21}. The statistical model makes it possible to perform a more robust probabilistic ``soft'' data association. In addition, heuristics for track initialization and termination can be avoided by modeling the existence of objects by binary random variables.

 \ac{bp} \cite{KscFreLoe:01} can solve high-dimensional Bayesian estimation problems by ``passing messages'' on the edges of a factor graph \cite{KscFreLoe:01} that represents the underlying statistical model. By exploiting the structure of the graph, \ac{bp}-based \ac{mot} methods \cite{Wil:J15,MeyBraWilHla:J17,MeyKroWilLauHlaBraWin:J18,MeyWin:J20,MeyWil:J21}  are highly scalable. This makes it possible to generate and maintain a very large number of potential object tracks and, in turn, outperform existing \ac{mot} approaches \cite{MeyBraWilHla:J17, MeyKroWilLauHlaBraWin:J18}.  However, \ac{bp} fully relies on a statistical model. When the factor graph does not accurately represent the true data generating process, \ac{mot} performance is reduced due to model mismatch. In addition, with the detect-then-track approach employed by \ac{bp}-based \ac{mot}, important object-related information might be discarded by the object detector. We aim to overcome these limitations by introducing information learned from raw sensor data to \ac{bp}-based \ac{mot}.

\ac{nebp} \cite{SatWel:21} is a hybrid method that combines the benefits of model-based and data-driven inference and addresses potential limitations of \ac{bp} such as model mismatch and overconfident beliefs \cite{SatWel:21,LiaMey:21}. \ac{nebp} has been successfully applied to decoding \cite{SatWel:21} and cooperative localization \cite{LiaMey:21} tasks. In \ac{nebp}, a \ac{gnn} that matches the topology of the factor graph is introduced. The trained \ac{gnn} enhances potentially inaccurate \ac{bp} messages to ultimately improve object declaration and estimation accuracy.  

In this paper, we propose \ac{nebp}  for \ac{mot}. Here, BP messages calculated as input for probabilistic data association are combined with the output of the \ac{gnn}. The \ac{gnn} uses measurements (i.e.,~object detections) and shape features learned from raw sensor data as an input. For \textit{false alarm rejection}, the \ac{gnn} identifies which measurements are likely false alarms. If a measurement has been identified as a potential false alarm, the false alarm distribution in the statistical model of \ac{bp} is locally increased. This reduces the probability that the measurement is associated with an existing object track or initializes a new object track. \textit{Object shape association} computes improved association probabilities by also comparing shape features extracted for existing object tracks with shape features extracted for measurements. The resulting \ac{nebp} method for \ac{mot} can improve object declaration and estimation performance compared to \ac{bp} for \ac{mot} as well as outperform further state-of-the-art methods. 

The main contributions of this paper can be summarized as follows.
\begin{itemize}[leftmargin=*]
    \item We introduce an \ac{nebp} method for \ac{mot} where probabilistic data association is augmented by shape features learned from raw sensor\vspace{1mm} data. 
    \item We apply the proposed method to the nuScenes autonomous driving dataset \cite{CaeBanLanVorLioXuKriPanBalBei:20} and demonstrate state-of-the-art object tracking performance.
\end{itemize}
Our approach recognizes that in modern MOT problems with high-resolution sensors \cite{GraBauReu:J17,GraFatSve:J19,GarWilSveXia:J21,MeyWil:J21}, it is challenging to capture object shapes and the corresponding data generating process by a statistical model. Consequently, the influence of object shapes on data generation is best learned directly from data\vspace{-2mm}. 

\section{Background}
In what follows, we will briefly review factor graphs and graph neural networks (GNNs).

\subsection{Factor Graphs}

A factor graph \cite{KscFreLoe:01} is a bipartite undirected graph $\Set{G}_f \rmv=\rmv (\Set{V}_f, \Set{E}_f)$ that consists of a set of edges $\Set{E}_f$ and a set of vertices or nodes $\Set{V}_f = \Set{Q} \cup \Set{F}$. A variable node $q \in \Set{Q}$ represents a random variable $\V{x}_q$ and a factor node $s \in \Set{F}$ represents a factor $f_s\big(\V{x}^{(s)}\big)$. Here, each factor argument $\V{x}^{(s)}$ comprises certain random variables $\V{x}_q$ (each $\V{x}_q$ can appear in several $\V{x}^{(s)}$). Factor nodes and variable nodes are typically depicted by circles and boxes, respectively. The joint \ac{pdf} represented by the factor graph reads $p(\V{x}) \propto \prod_{s \in \Set{F}} f_s\big(\V{x}^{(s)}\big)$ where $\propto$ indicates equality up to a constant factor.  

Belief propagation \acused{bp} (\ac{bp}) \cite{KscFreLoe:01}, also known as the sum-product algorithm can compute marginal \acp{pdf} $p(\V{x}_q)$, $q \rmv\in\rmv \Set{Q} $ efficiently. \ac{bp} performs local operations called ``messages'' on the edges of the factor graph. There are two types of messages: 
(i) messages passed from variable nodes to factor nodes given\vspace{-.5mm} by 
\begin{equation}
\phi_{\V{x}_q \to f_s}(\V{x}_q) = \prod_{a \in \Set{N}_{\Set{F}}(q) \backslash s} \phi_{f_a \to \V{x}_q}(\V{x}_q) \label{eq:bp_x_to_f} \nn
\vspace{.5mm}
\end{equation}
and (ii) messages passed from factor nodes to variable nodes defined as
\begin{equation}
\phi_{f_s \to \V{x}_q}(\V{x}_q) = \sum_{\V{x}^{(s)} \backslash \V{x}_q} f_s\big(\V{x}^{(s)}\big) \prod_{m \in \Set{N}_{\Set{Q}}(s) \backslash q} \phi_{\V{x}_m \to f_s}(\V{x}_m). \nn
\vspace{.5mm}
\end{equation}
Here, $\Set{N}_{\Set{Q}}(\cdot) \rmv\subseteq\rmv \Set{Q}$ and $\Set{N}_{\Set{F}}(\cdot) \rmv\subseteq\rmv \Set{F}$ denote the set of neighboring variable and factor nodes, respectively. After message passing is completed, one can subsequently obtain marginal \acp{pdf} $p(\V{x}_q)$ as the product of all incoming messages from the neighboring factors, i.e.,  $p(\V{x}_q) \propto \prod_{s \in \Set{N}_{\Set{F}}(q)} \phi_{f_s \to \V{x}_q}(\V{x}_q)$. In factor graph with loops, \ac{bp} is applied in an iterative manner. It can then only provide approximations of marginal posterior pdfs $p(\V{x}_q)$. 
\vspace{.7mm}

\subsection{Graph Neural Networks}
Graph Neural Networks \acused{gnn} (\acp{gnn}) \cite{GorMonSca:05} extend neural networks to graph-structured data. We consider \acp{mpnn} \cite{GilSchRilVinDah:17} which are a variant of \acp{gnn} that generalizes graph convolutional networks \cite{KipWel:17} and provides a message passing mechanism similar to \ac{bp}. A  \ac{mpnn} is defined on a graph $\Set{G} = (\Set{V}, \Set{E})$ where $\Set{E}$ induces the sets of neighbors $\Set{N}(i) = \{j \in \Set{V} \big| (i,j) \in \Set{E} \}$.

Each node $i \in \Set{V}$ is associated with a vector $\V{h}_i$ called node embedding. At message passing iteration $l \in \{1,\dots,L\}$, the following operations are performed for each node $i \in \Set{V}$ in parallel. First, messages are exchanged with neighboring nodes $j \rmv\in \Set{N}(i)$. In particular, the \ac{gnn} message sent from node $i \in \Set{V}$ to its neighboring node $j \in \Set{N}(i)$ is given\vspace{1mm} by 
\begin{equation}
\V{m}_{i \to j}^{(l)} = g_{i \to j} \big(\V{h}_{i}^{(l)}\rmv\rmv, \V{h}_{j}^{(l )}\rmv\rmv, \V{a}_{i \to j}\big). \nn
\vspace{1mm}
\end{equation}
In addition, the node embedding $\V{h}^{(l)}_i$ is updated by incorporating the sum of received\vspace{.5mm} messages 
$\V{m}_{j \to i}^{(l)}$, $j \in \Set{N}(i)$, i.e., 
\begin{equation}
\V{h}_{i}^{(l + 1)} = g_{i} \Big(\V{h}_{i}^{(l)}\rmv\rmv, \sum_{j \in \Set{N}(i)} \V{m}_{j \to i}^{(l)}, \V{a}_i \Big). \nn
\vspace{.5mm}
\end{equation}
Here, $g_{i} (\cdot), i \in \Set{V}$ and $g_{i \to j}(\cdot), (i, j) \in \Set{E}$ denote the node and edge networks, respectively. Furthermore, $\V{a}_i$ and $\V{a}_{i \to j}$ denote node and edge\vspace{0mm} attributes.

\begin{figure*}
    \centering
    \subfloat{
	\psfrag{oa1}[c][c][0.8]{\color{myblue}{\raisebox{2mm}{\hspace{2mm}{$\kappa_{1}$}}}}
	\psfrag{ob1}[c][c][0.8]{\color{myblue}{\raisebox{2.5mm}{\hspace{2mm}{$\iota_{1}$}}}}
    \psfrag{da1}[c][c][0.8]{\raisebox{-2mm}{\hspace{.1mm}$a_{1}$}}
    \psfrag{daI}[c][c][0.8]{\raisebox{-2mm}{$a_{I}$}}
    \psfrag{db1}[c][c][0.8]{\hspace{.5mm}$b_{1}$}
    \psfrag{dbJ}[c][c][0.8]{\hspace{.4mm}\raisebox{1mm}{$b_{J}$}}
    \psfrag{g11}[c][c][0.7]{{\hspace{.42mm}\raisebox{-3mm}{$\Psi_{1,1}$}}}
    \psfrag{g1I}[c][c][0.7]{{\hspace{-.3mm}\raisebox{-3.5mm}{$\Psi_{1,J}$}}}
    \psfrag{gJ1}[c][c][0.7]{{\hspace{-.4mm}\raisebox{-3mm}{$\Psi_{I,1}$}}}
    \psfrag{gIJ}[c][c][0.7]{{\hspace{-1.1mm}\raisebox{-3mm}{$\Psi_{I,J}$}}}
	\psfrag{ma1}[c][c][0.8]{\color{myblue}{$\phi_{a_1}$}}
	\psfrag{mb1}[c][c][0.8]{\color{myblue}{$\phi_{b_1}$}}
	\psfrag{mab11}[c][c][0.8]{\color{myblue}{\hspace{1mm}{$\phi_{\Psi_{1,1} \to b_1}$}}}
	\psfrag{mbaJ1}[c][c][0.8]{\color{myblue}{\hspace{-1mm}{$\phi_{\Psi_{1,J} \to a_1}$}}}
	\psfrag{q1}[c][c][0.8]{\hspace{.4mm}\raisebox{0mm}{$q_1$}}
	\psfrag{qI}[c][c][0.8]{\hspace{.4mm}\raisebox{-2mm}{$q_I$}}
	\psfrag{v1}[c][c][0.8]{\raisebox{-2.3mm}{\hspace{.5mm}\hspace{.2mm}\raisebox{-.5mm}{$v_1$}}}
	\psfrag{vJ}[c][c][0.8]{\raisebox{-1.8mm}{\hspace{.5mm}\hspace{-.5mm}\raisebox{-1.2mm}{$v_J$}}}
	\psfrag{yu1}[c][c][0.8]{\hspace{.6mm}\raisebox{-2.3mm}{$\underline{\V{y}}{}_1$}}
	\psfrag{yuI}[c][c][0.8]{\hspace{.5mm}\raisebox{-2.3mm}{$\underline{\V{y}}{}_I$}}
	\psfrag{yo1}[c][c][0.8]{\hspace{.65mm}$\overline{\V{y}}{}_1$}
	\psfrag{yoJ}[c][c][0.8]{\hspace{.65mm}$\overline{\V{y}}{}_J$}
	\psfrag{f1}[c][c][0.8]{$f_1$}
	\psfrag{fI}[c][c][0.8]{\hspace{.3mm}$f_I$}
	\psfrag{L2}[c][c][1]{\hspace{.65mm}(a)}
    \hspace{-10mm}\includegraphics[scale=0.8]{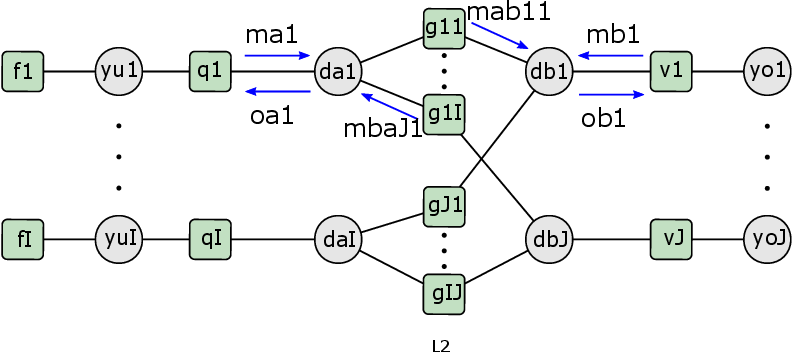} \label{fig:fg}}
    \hspace{13mm}
    \subfloat{\raisebox{9.65mm}{
    \psfrag{da1}[c][c][0.8]{\hspace{1mm}\raisebox{-2.5mm}{$\V{h}_{a_{1}}$}}
    \psfrag{dan}[c][c][0.8]{\hspace{1mm}$\V{h}_{a_{I}}$}
    \psfrag{db1}[c][c][0.8]{\hspace{.8mm}\raisebox{-2.5mm}{$\V{h}_{b_{1}}$}}
    \psfrag{dbm}[c][c][0.8]{\hspace{1mm}\raisebox{-2.5mm}{$\V{h}_{b_{J}}$}}
	\psfrag{mab11}[c][c][0.8]{\hspace{-.5mm}\raisebox{-3.5mm}{\rd{$\V{m}_{a_1 \to b_1}$}}}
	\psfrag{L1}[c][c][1]{\raisebox{-21.5mm}{\hspace{.65mm}(b)}} 
    \includegraphics[scale=0.8]{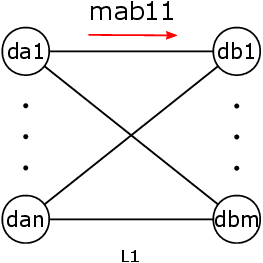}} \label{fig:gnn}}
    \captionsetup{singlelinecheck = false, justification=justified}	
    \vspace*{1mm}
    \caption{ Factor graph for \ac{mot} (a) and \vspace{-.5mm} corresponding \acr{gnn} (b) for a\vspace{.7mm} single time frame $k$. In (a), \ac{bp} messages that correspond to \ac{da} are shown in blue color. These messages are enhanced by the proposed \ac{nebp} approach are shown; (b) shows the corresponding \ac{gnn} messages. The time index $k$ is omitted.}
    \label{fig:fg_gnn}
    \vspace{-1mm}
\end{figure*}

\section{Review of BP-based Multi-Object Tracking}

The proposed \ac{nebp} approach is based on \ac{bp}-based \ac{mot} introduced in \cite{MeyKroWilLauHlaBraWin:J18}. The statistical model used by \ac{bp}-based \ac{mot} is reviewed next. 

\subsection{Object States}

At each time frame $k$, an object detector $g_{\text{det}}(\cdot)$ extracts $J_k$ measurements $\V{z}_{k} \triangleq [\V{z}_{k, 1}^\T \cdots \V{z}_{k, J_k}^\T]^\T$ from raw sensor data $\Set{Z}_k$, i.e., $\V{z}_{k} = g_{\text{det}}(\Set{Z}_k)$. All measurements extracted up to time frame $k$ are denoted as $\V{z}_{1 : k} \triangleq [\V{z}_{1}^\T \cdots \V{z}_{k}^\T]^\T\rmv$.
Since the number objects is unknown, \ac{po} states are introduced. The number of \ac{po} states $N_k$ is the maximum possible number of objects that have generated a measurement up to time frame $k$. At time frame $k$, the existence of a \ac{po} $n \in \{1, \cdots, N_k\}$ is modeled by a binary random existence variable $r_{k, n} \in \{0, 1\}$, i.e., \ac{po} $n$ exists if and only if $r_{k, n} = 1$. The state of \ac{po} $n$ is modeled by the random vector $\V{x}_{k, n}$. The augmented \ac{po} state\vspace{-.5mm} vector is denoted by $\V{y}_{k, n} \triangleq [\V{x}_{k, n}^\T \hspace{1mm} r_{k, n}]^\T$ and the joint \ac{po} state vector\vspace{-.3mm} by $\V{y}_{k} \triangleq [\V{y}_{k, 1}^\T \cdots \V{y}_{k, N_k}^\T]^\T\rmv\rmv$. There are two types of \acp{po}:
\begin{itemize}[leftmargin=*]
    \item \textit{New \acp{po}} denoted $\overline{\V{y}}{}_{k, j} = [\overline{\V{x}}{}_{k, j}^\T \hspace{1mm} \overline{r}{}_{k, j}]^\T\rmv$, $j \in \{1, \cdots, J_k\}$ represent objects that at time frame $k$ generated a measurement for the first time. Each measurement $\V{z}_{k, j}, j \rmv\in\rmv \{1, \cdots, J_k\}$ introduces a new \ac{po} $j$ with state $\overline{\V{y}}{}_{k, j}$. 
    \item \textit{Legacy \acp{po}} denoted $\underline{\V{y}}{}_{k, i} = [\underline{\V{x}}{}_{k, i}^\T \hspace{1mm} \underline{r}{}_{k, i}]^\T\rmv$, $i = \{1, \cdots, I_k\}$ represent objects that have generated a measurement for the first time at a previous time frame $k^\prime < k$.
\end{itemize}

New \acp{po} become legacy \acp{po} when the measurements of the next time frame are considered. Thus, the number of legacy \acp{po} at time frame $k$ is $I_k \rmv=\rmv I_{k - 1} \rmv+\rmv J_{k - 1} \rmv=\rmv N_{k - 1}$ and the total number of \acp{po} is $N_k \rmv=\rmv I_k \rmv+\rmv J_k$. We further denote the joint new \ac{po} state by $\overline{\V{y}}{}_{k} \rmv\triangleq\rmv [\overline{\V{y}}{}_{k, 1}^\T \cdots \overline{\V{y}}{}_{k, J_k}^\T ]^\T$ and the joint legacy \ac{po} state by $\underline{\V{y}}{}_{k} \rmv\triangleq\rmv [\underline{\V{y}}{}_{k, 1}^\T \cdots \underline{\V{y}}{}_{k, I_k}^\T ]^\T\rmv\rmv$, i.e., $\V{y}_k = [\underline{\V{y}}{}_{k}^\T \hspace{1mm} \overline{\V{y}}{}_{k}^\T]^\T\rmv$.
\vspace{.8mm}

\subsection{Measurement Model}

The origin of measurements $\V{z}_{k, j}$, $j \in \{1, \cdots, J_k\}$ is unknown. A measurement can originate from a \ac{po} or can be a false alarm. Furthermore, a \ac{po} may also not generate any measurements (missed detection). With the assumption that a \ac{po} can generate at most one measurement and a measurement is originated from at most one \ac{po}, we model data association uncertainty as follows \cite{MeyKroWilLauHlaBraWin:J18}. The \ac{po}-measurement association at time frame $k$ can be described by an ``object-oriented'' \ac{da} vector $\V{a}_k \rmv=\rmv [a_{k, 1} \cdots a_{k, I_k}]^\T$. Here, the association variable $a_{k, i} \rmv=\rmv j \in \{1, \cdots, J_k\}$ indicates that legacy \ac{po} $i$ generated measurement $j$ and $a_{k, i} = 0$ indicated that legacy \ac{po} $i$ did not generate any measurement at time $k$. Following \cite{WilLau:J14}, we also introduce the ``measurement-oriented'' \ac{da} vector $\V{b}_k = [b_{k, 1} \cdots b_{k, J_k}]^\T$ with $b_{k, j} \rmv=\rmv i \in \{1, \cdots, I_k\}$ if measurement $j$ was generated by legacy \ac{po} $i$, or $b_{k, j} = 0$ if measurement $j$ was not generated by any legacy \ac{po}. Note that there is a one-to-one mapping between $\V{a}_k$ and $\V{b}_{k}$ and vice versa. Introducing $\V{b}_{k}$ in addition to $\V{a}_k$ makes it possible to develop scalable \ac{mot} \cite{MeyKroWilLauHlaBraWin:J18}. Finally,\vspace{0mm} we establish the notation $\V{a}_{1 : k} \triangleq [\V{a}_1^\T \cdots \V{a}_{k}^\T]^\T$ and $\V{b}_{1 : k} \triangleq [\V{b}_1^\T \cdots \V{b}_{k}^\T]^\T\rmv$.


 
If legacy \ac{po} $i$ exists, it generates a measurement (i.e. $a_{k, i} \rmv=\rmv j \in \{1, \cdots, J_k\}$) with probability $p_{\text{d}}$. Furthermore, the probability that it also exists at the next time frame $k\rmv+\rmv1$ is $p_{\text{s}}$. The number of false alarms is modeled by a Poisson distribution with mean $\mu_{\text{fa}}$ and false alarm measurements are independent and identically distributed according to $f_{\text{fa}} (\V{z}_{k, j})$. Before the measurements $\{\V{z}_{k, j}\}_{j = 1}^{J_k}$ are observed, the number of new \acp{po} is unknown. The number of newly detected objects is Poisson distributed with mean $\mu_{\text{n}}$, while the states of newly detected objects are a priori independent and identically distributed according to $f_{\text{n}}(\overline{\V{x}}{}_{k, j})$. Following the assumptions presented in \cite[Sec. VIII-A]{MeyKroWilLauHlaBraWin:J18}, the joint posterior \ac{pdf} $f(\V{y}_{1 : k}, \V{a}_{1 : k}, \V{b}_{1 : k} | \V{z}_{1 : k})$ can be derived \cite[Sec. VIII-G]{MeyKroWilLauHlaBraWin:J18}. The factorization of this joint posterior pdf is visualized by the factor graph shown in Fig.~\ref{fig:fg}. Note that legacy POs are connected to object-oriented association variables and new POs are connected to measurement-oriented association variables.

\subsection{Object Declaration and State Estimation}

In the Bayesian setting, declaration of object existence and object state estimation are based on the marginal existence probabilities $p(r_{k, n} \rmv=\rmv 1 | \V{z}_{1 : k})$ and the conditional \acp{pdf} $f(\V{x}_{k, n} | r_{k, n} \rmv=\rmv 1, \V{z}_{1 : k})$. In particular, declaration of object existence is performed by comparing $p(r_{k, n} \rmv=\rmv 1 | \V{z}_{1 : k})$ to a threshold  $T_{\text{dec}}$. In addition, for objects $n$ that are declared to exist, an estimate of $\V{x}_{k, n}$ is provided by the \ac{mmse}\vspace{.9mm} estimator
\begin{equation}
 \hat{\V{x}}{}_{k, n}^{\text{MMSE}} \rmv=\rmv \int \V{x}_{k, n} f(\V{x}_{k, n} \ist|\ist r_{k, n} \rmv=\rmv 1, \V{z}_{1 : k}) \ist \mathrm{d}\V{x}_{k, n}. \nn
 \vspace{.9mm}
 \end{equation} 
 Note that declaration of object existence is based on the existence probability $p(r_{k, n} \rmv= 1 \ist | \ist \V{z}_{1 : k}) \rmv=\rmv \int f(\V{x}_{k, n}, r_{k, n} \rmv=\rmv 1 \ist |$ $\V{z}_{1 : k}) \mathrm{d}\V{x}_{k, n}$. In addition, object state estimation makes use of
 \vspace{-1mm} 
 \begin{equation}
f(\V{x}_{k, n} \ist | \ist r_{k, n} \rmv=\rmv 1, \V{z}_{1 : k}) \rmv=\rmv \frac{f(\V{y}_{k, n} \ist | \ist \V{z}_{1 : k})}{p(r_{k, n} = 1 \ist | \ist \V{z}_{1 : k})}. \nn
\vspace{.8mm}
\end{equation}
Thus, both tasks rely on the calculation of marginal posterior \acp{pdf} $f(\V{y}_{k, n} \ist | \ist \V{z}_{1 : k}) \rmv\triangleq\rmv f(\V{x}_{k, n}, r_{k, n} \ist | \ist \V{z}_{1 : k})\vspace{0mm}$.  By applying \ac{bp} following \cite[Sec. VIII-IX]{MeyKroWilLauHlaBraWin:J18}, accurate approximations (a.k.a. ``beliefs'') $\tilde{f}(\V{y}_{k, n}) \rmv\approx\rmv f(\V{y}_{k, n} | \V{z}_{1 : k})\vspace{-.5mm}$ of marginal posterior \acp{pdf} can be calculated efficiently. For future reference we introduce the notation $\hat{r}_{k, n} = p(r_{k, n} \rmv= 1 \ist | \ist \V{z}_{1 : k})$.

Note that since we introduce a new \ac{po} for each measurement, the number of POs grows with time $k$. Therefore, legacy and new POs whose approximate existence probabilities are below a threshold $T_{\text{pru}}$ are pruned, i.e., removed from the state\vspace{-.5mm} space.

\section{NEBP-based Multi-Object Tracking}
\vspace{-.3mm}

To further improve the performance of \ac{bp}-based \ac{mot}, we augment the factor graph in Fig.~\ref{fig:fg} by a \ac{gnn}. The \ac{gnn} uses features extracted from previous estimates, measurements, and raw sensor information as an input. Since we limit the following discussion to a single time frame, we will omit the time index $k$.

\subsection{Feature Extraction} \label{sec:feaExtraction}
First, we discuss how features are learned from raw sensor data for legacy \acp{po} and measurements. We consider motion and shape features. The motion features for legacy \ac{po} $i$ and measurement $j$ are computed as $\V{h}_{a_i, \text{motion}} = g_{\text{motion}, 1}(\underline{\hat{\V{x}}}{}_{i}^{-}, \underline{\hat{r}}{}_{i}^{-})$ and $\V{h}_{b_j, \text{motion}} = g_{\text{motion}, 2}(\V{z}_{j})$, respectively. Here, $g_{\text{motion}, 1}(\cdot)$ as well as $g_{\text{motion}, 2}(\cdot)$ are neural networks.  
In addition, $\underline{\hat{\V{x}}}{}_{i}^{-}$ and $\underline{\hat{r}}{}_{i}^{-}$ are the approximate \ac{mmse} state estimate and existence probability of legacy \ac{po} $i$ at the previous time frame. Similarly, the shape features, denoted by $\V{h}_{a_i, \text{shape}}$ and $\V{h}_{b_j, \text{shape}}$, are extracted from raw sensor data $\Set{Z}^{-}$ and $\Set{Z}$ at previous and current time, respectively, i.e., $\V{h}_{a_i, \text{shape}} = g_{\text{shape}, 1}(\Set{Z}^{-}\rmv\rmv, \underline{\hat{\V{x}}}{}_{i}^{-})$ and $\V{h}_{b_j, \text{shape}} = g_{\text{shape}, 2}(\Set{Z},\V{z}_{j})$. Here, $g_{\text{shape}, 1}(\cdot)$ as well as $g_{\text{shape}, 2}(\cdot)$ are again neural networks. We will discuss one particular instance of shape feature extraction in\vspace{0mm} Sec.~\ref{sec:exp_setup}. 

\subsection{The Proposed Message Passing Algorithm}

For neural enhanced \ac{da}, we introduce a \ac{gnn} that matches the topology of the \ac{da} section of the factor graph in Fig.~\ref{fig:fg}. The resulting \ac{gnn} is shown in Fig.~\ref{fig:gnn}.  In addition to the output of the detector, the \ac{gnn} also uses raw sensor information as an input. The goal is to use this additional information to reject false alarm measurements and obtain improved \ac{da} probabilities by enhancing \ac{bp} messages with the output of the \ac{gnn}. 
\ac{nebp} for \ac{mot} consists of the following\vspace{.5mm} steps:

\subsubsection{Conventional \ac{bp}}

First, conventional \ac{bp}-based \ac{mot} is run until convergence. This results in the \ac{bp} messages $\phi_{a_i} = [\phi_{a_i}(0) \cdots \phi_{a_i}(J)]^\T \in \mathbb{R}^{J + 1}, \phi_{b_j} = [\phi_{b_j}(0) \cdots \phi_{b_j}(I)]^\T \in \mathbb{R}^{I + 1}, \phi_{\Psi_{i, j} \to b_j} \in \mathbb{R}$, and $\phi_{\Psi_{i, j} \to a_i} \in \mathbb{R}$ (cf.~\cite[Sec.~IX-A1--IX-A3]{MeyKroWilLauHlaBraWin:J18}).
\vspace{1.5mm}

\subsubsection{\ac{gnn} Messages}

 Next, \ac{gnn} message passing is\vspace{.5mm} executed iteratively. In particular, at iteration $l \rmv\in\rmv \{1,\dots,L\}$ the following operations are performed:
    \begin{align}
    	\V{m}_{a_i \to b_j}^{(l)} &= g_{\text{e}} \Big( \V{h}_{a_i}^{(l)}, \V{h}_{b_j}^{(l)}, \phi_{a_i}(j), \phi_{\Psi_{i, j} \to b_j} \Big) \label{eq:gnn_a_to_b} \\[1.2mm]
    	\V{m}_{b_j \to a_i}^{(l)} &= g_{\text{e}} \Big(\V{h}_{a_i}^{(l)}, \V{h}_{b_j}^{(l)}, \phi_{a_i}(j), \phi_{\Psi_{i, j} \to a_i} \Big) \nn\\[1.2mm]
    	\V{h}_{a_i}^{(l+1)} &= g_{\text{n}} \bigg(\V{h}_{a_i}^{(l)}, \sum_{j \in \mathcal{N}(i)} \V{m}_{b_j \to a_i}^{(l)}, \phi_{a_i}(0) \bigg) \nn\\[.8mm]
    	\V{h}_{b_j}^{(l+1)} &= g_{\text{n}} \bigg(\V{h}_{b_j}^{(l)}, \sum_{j \in \mathcal{N}(i)} \V{m}_{a_i \to b_j}^{(l)}, \phi_{b_j}(0) \bigg). \label{eq:gnn_b}
    \end{align}
    Here, $g_{\text{e}}(\cdot)$ is the edge neural network and $g_{\text{n}}(\cdot)$ is the node neural network. The edge neural network $g_{\text{e}}(\cdot)$ provides messages passed along the edges of the \ac{gnn}.
    
    The node embeddings are initialized as the concatenation of respective motion and shape features\vspace{-.3mm}, i.e., $\V{h}_{a_i}^{(1)} \rmv=\rmv [\V{h}_{a_i, \text{motion}}^\T$ $\V{h}_{a_i, \text{shape}}^\T]^\T$ and $\V{h}_{b_j}^{(1)} \rmv=\rmv [\V{h}_{b_j, \text{motion}}^\T$ $\V{h}_{b_j, \text{shape}}^\T]^\T\rmv\vspace{0mm}$. Finally\vspace{-.5mm}, for each $j \in \{1, \cdots, J\}$, the correction\vspace{-1mm} factors $\beta_{j} = g_{\text{r}}(\V{h}_{b_j}^{(L)}) \in (0, 1]$ and $\gamma_{i}(j) = g_{\text{a}}\big(\V{m}_{b_j \to a_i}^{(L)}\big)$ $\in \mathbb{R}$ are computed based\vspace{-.5mm} on the two additional neural networks $g_{\text{r}}(\cdot)$ and $g_{\text{a}}(\cdot)$. As discussed next, these correction factors provided by the \ac{gnn} are used to implement  false alarm rejection and object shape association, respectively\vspace{1mm}.
    
    \subsubsection{NEBP Messages}
     After computing \eqref{eq:gnn_a_to_b}--\eqref{eq:gnn_b} for $L$ iterations, neural enhanced message passing is performed as\vspace{0mm} follows. First, neural enhanced versions $\tilde{\phi}_{a_i}$ of the messages $\phi_{a_i}$ are\vspace{0mm} obtained by computing\vspace{.5mm}
    \begin{equation}
        \tilde{\phi}_{a_i}(j) = \beta_{j} \bar{\phi}_{a_i}(j) + \text{ReLU}\big(\gamma_{i}(j) \big), \quad j \in \{1, \cdots, J\}
        \label{eq:nebp_combine_a}
        \vspace{.5mm}
    \end{equation}
    and setting $\tilde{\phi}_{a_i}(0) \rmv=\rmv \phi_{a_i}(0)$. Here, $\text{ReLU}(\cdot)$ is a rectified linear unit and $\bar{\phi}_{a_i}$ is a normalized\footnote{Multiplying BP messages by a constant factor does not alter the resulting beliefs \cite{KscFreLoe:01}.} version of $\phi_{a_i}$ (cf.~\cite[Sec.~IX-A2]{MeyKroWilLauHlaBraWin:J18}),\vspace{-1mm} i.e.,
  \begin{equation}  
\bar{\phi}_{a_i}(j) = \frac{\phi_{a_i}(j)}{\sum_{j^\prime = 0}^{J}\phi_{a_i}(j^\prime)}, \quad j \in \{1, \cdots, J\}. \nn
\vspace{0mm}
  \end{equation}
Note that $\phi_{a_i}(j)$, $j \rmv\in\rmv \{1,\dots,J\}$ represents the likelihood that the legacy \ac{po} $i$ is associated to measurement $j$ \cite{MeyKroWilLauHlaBraWin:J18}. Consequently, the $\text{ReLU}\big(\gamma_{i}(j) \big) \rmv>\rmv 0$ term in \eqref{eq:nebp_combine_a} provided by the \ac{gnn} implements object shape association, i.e., the likelihood that the legacy \ac{po} $i$ is associated to the measurement $j$ is increased if the shape features extracted for legacy POs resembles the shape features extracted for measurements.

   Next, neural enhanced versions $\tilde{\phi}_{b_j}$ of the messages $\phi_{b_j}$ are obtained by computing
	\begin{equation}
        \tilde{\phi}_{b_j}\rmv(0) = \beta_{j} \big(\phi_{b_j}(0) - 1\big) + 1
        \label{eq:nebp_combine_b}
        \vspace{.5mm}
    \end{equation}
    and setting $\tilde{\phi}_{b_j}(i) = \phi_{b_j}(i), i \in \{1, \cdots, I\}$. We recall that $\phi_{b_j}(0)$ is given by (cf.~\cite[Sec.~IX-A2]  {MeyKroWilLauHlaBraWin:J18})
     \begin{align}
     \phi_{b_j}\rmv(0) =  \frac{\mu_{\text{n}} }{\mu_{\text{fa}} \ist f_{\text{fa}}\big( \V{z}_{j} \big) } \int \! f_{\text{n}}\big( \overline{\V{x}}{}_{j} \big)\ist f\big( \V{z}_{j} \big| \overline{\V{x}}{}_{j} \big) \ist \mathrm{d}\overline{\V{x}}{}_{j} + 1\ist. \nn\\[-4mm]
\nn
     \end{align}
     
     The scalar $\beta_{j} \rmv\rmv\in\rmv\rmv (0, 1) $ in \eqref{eq:nebp_combine_a} and \eqref{eq:nebp_combine_b} provided by the \ac{gnn} implements false alarm rejection. In particular, $\beta_{j} < 1$ is equal to the local increase of the false alarm distribution given by $\tilde{f}_{\text{fa}}(\V{z}_{j}) = \frac{1}{\beta_{j}} f_{\text{fa}}(\V{z}_{j})$. This local increase of the false alarm distribution makes it less likely that the measurement $\V{z}_{j}$ is associated to a legacy PO and reduces the existence probability of the  new PO introduced for the measurement $\V{z}_{j}$ \vspace{1mm}.
    
    
     \subsubsection{Belief Calculation}
Finally, conventional \ac{bp}-based \ac{mot} is again run until convergence by\vspace{.3mm} replacing $\phi_{a_i}$ with its neural enhanced counterpart $\tilde{\phi}_{a_i}$. This results in the enhanced output messages  $\tilde{\kappa}_{i} = [\tilde{\kappa}_{i}(0) \cdots \tilde{\kappa}_{i}(J)]^\T \in \mathbb{R}^{J + 1}$ and $\tilde{\iota}_{j} = [\tilde{\iota}_{i}(0) \cdots \tilde{\iota}_{i}(I)]^\T \in \mathbb{R}^{I + 1}$ (cf. Fig.~\ref{fig:fg_gnn}). After performing the normalization\vspace{-.5mm}  
    \begin{equation}
    \tilde{\kappa}^\prime_{i}(j) = \frac{\tilde{\phi}_{a_i}(j)}{\phi_{a_i}(j)} \tilde{\kappa}_{i}(j), \quad j \in \{0, \cdots, J\} \nn
    \vspace{.5mm}
    \end{equation}    
the resulting\vspace{.2mm} messages $\tilde{\kappa}_i^\prime$ are used for the calculation of legacy \ac{po} beliefs $\tilde{f}(\underline{\V{y}}{}_{i})$, $i \in \{1, \cdots, I\}$ (cf.~\cite[Sec.~IX-A4--IX-A6]{MeyKroWilLauHlaBraWin:J18}). Similarly, the enhanced messages $\tilde{\iota}_{j}$ are directly used for the calculation of new \ac{po} beliefs $\tilde{f}(\overline{\V{y}}{}_{j}), j \in \{1, \cdots, J\}$.

\subsection{The Loss Function} \label{sec:Loss}

For supervised learning, it is assumed that ground truth object tracks are available in the training set. Ground truth object tracks consist of a sequence of object positions and object identities (IDs). During the training of the \ac{gnn}, the parameters of all neural networks are updated through back-propagation, which computes the gradient of the loss function. The loss function has the form $\Set{L} = \Set{L}_{\text{r}} + \Set{L}_{\text{a}}$. Here, the two contributions $\Set{L}_{\text{r}}$ and $\Set{L}_{\text{a}}$ establish false alarm rejection and object shape association, respectively. 

False alarm rejection, introduces the binary cross-entropy loss \cite[Chapter 4.3]{Bishop:B06}\vspace{0mm}
\begin{equation}
    \Set{L}_{\text{r}} = \frac{-1}{J} \sum_{j = 1}^{J}  \beta_j^{\text{gt}} \ln(\beta_{j}) + \epsilon (1 - \beta_j^{\text{gt}}) \ln(1 - \beta_{j}) \label{eq:loss_beta} \vspace{0mm}
\end{equation}
where $\beta_j^{\text{gt}} \rmv\in\rmv \{0,1\}$ is the pseudo ground truth label for each measurement and $\epsilon \rmv\in\rmv \mathbb{R}^{+}$ is a tuning parameter. $\beta_j^{\text{gt}}$ is $1$ if the distance between the measurement and any ground truth position is not larger than $T_{\text{dist}}$ and $0$ otherwise. 

The tuning parameter $\epsilon \rmv\in\rmv \mathbb{R}^{+}$ is motivated as follows. Since missing an object is typically more severe than producing a false alarm, object detectors often output many detections and produce more false alarm measurements than true measurements. The tuning parameter $\epsilon \rmv\in\rmv \mathbb{R}^{+}$ addresses this imbalance problem which is well studied in the context of learning-based binary classification \cite{OksCamKalAkb:20}.

Since  $\tilde{\phi}_{a_i}(j)$ in \eqref{eq:nebp_combine_a} represents the likelihood that the legacy \ac{po} $i$ is associated to the measurement $j$, ideally $\text{ReLU}\big(\gamma_{i}(j) \big)$ is large if \ac{po} $i$ is associated to the measurement $j$, and is equal to zero if they are not associated. Thus, object shape association introduces the following binary cross-entropy\vspace{1.5mm} loss 
\begin{align}
    \Set{L}_{\text{a}} = \frac{-1}{IJ} \sum_{i = 1}^{I} \sum_{j = 1}^{J}  &\gamma^{\text{gt}}_{i}(j) \ln \big( \sigma(\gamma_{i}(j)) \big) \nn \\[0.5mm]
    &+ \big( 1 - \gamma^{\text{gt}}_{i}(j) \big) \ln \big( 1 - \sigma(\gamma_{i}(j)) \big) \vspace{0mm} \label{eq:loss}\\[-2mm]
    \nn
\end{align}
where $\sigma(x) = 1/(1 + e^{-x})$ is the sigmoid function and $\gamma^{\text{gt}}_{i} = [\gamma^{\text{gt}}_{i}(1) \cdots \gamma^{\text{gt}}_{i}(J)]^\T \in \{0,1\}^J$ is the pseudo ground truth association vector of legacy \ac{po} $i \rmv\in\rmv\{1,\dots,I\}$. In each pseudo ground truth association vector $\gamma^{\text{gt}}_{i}$, at most one element is equal to one and all the other elements are equal to zero. 

Note that in \eqref{eq:loss}, we do not apply the ReLU to the $\gamma_{i}(j)$, since this would result in the gradients $\partial \Set{L}_{\text{a}} / \partial \gamma_{i}(j)$ to be zero for negative values of $\gamma_{i}(j)$. It was observed that performing backpropagation by also making use of the gradients related to the negative values of $\gamma_{i}(j)$, leads to a more efficient training of the \ac{gnn}.
At each time frame, pseudo ground truth association vectors are constructed from measurements and ground truth object tracks based on the following\vspace{.5mm} rules:
\begin{itemize}[leftmargin=*]
    \item \textit{Get Measurement IDs:} Compute the Euclidean distance between all ground truth positions and measurements and run the Hungarian algorithm \cite{BarWilTia:B11} to find the best association between ground truth positions and measurements. All measurements that have been associated with a ground truth position and have a distance to that ground truth position that is smaller than $T_{\text{dist}}$ inherit the ID of the ground truth position. All other measurements do not have an ID. \vspace{1mm}
    \item \textit{Update Legacy PO IDs:} Legacy POs inherit the ID from the previous time frame. If a legacy PO with ID has a distance not larger than $T_{\text{dist}}$ to a ground truth position with the same ID, it keeps its ID. The for a legacy PO $i \rmv\in\rmv \{1,\dots,I\}$ that has the same ID as measurement $j \rmv\in\rmv \{1,\dots,J\}$, the entry $\gamma^{\text{gt}}_{i}(j)$ is set to one. All other entries $\gamma^{\text{gt}}_{i}(j)$, $i \rmv\in\rmv \{1,\dots,I\}$, $j \rmv\in\rmv \{1,\dots,J\}$ are set to\vspace{1mm} zero.
    \item \textit{Introduce New PO IDs:} For any measurement $j \rmv\in \{1,\dots,J\}$ with an ID that does not share its ID with a legacy object, the corresponding new PO inherits the ground truth ID from the measurement. All other new POs do not have an\vspace{-.5mm} ID.
\end{itemize}

\begin{table*}[!t]
    \normalsize
    \centering
    
    \begin{tabular}{c|c|cccc} \hline
    Methods                                & Modalities   & AMOTA $\uparrow$ & IDS $\downarrow$ & Frag $\downarrow$  \\ \hline \hline
    CenterPoint \cite{YinZhoKra:21}        & LiDAR        & 0.665            & 562              & 424                 \\
    Chiu et al. \cite{ChiLiAmbBoh:21}      & LiDAR+Camera & 0.687            & -                & -                           \\
    Zaech et al. \cite{ZaeDaiLinDanVan:22} & LiDAR        & 0.693            & 262              & 332                        \\ \hline
    BP                                     & LiDAR        & 0.698            & \textbf{161}     & \textbf{250}                        \\
    NEBP (proposed)                        & LiDAR        & \textbf{0.708}   & 172              & 271 \\ 
    \hline        
    \end{tabular}
    \vspace{1mm}
    \caption{Performance results on nuScenes validation set. ``-'' indicates that the metric is not reported. }
    \label{tab:result}
    \vspace{-1mm}
\end{table*}

\section{Experimental Results}

We present results in an urban autonomous driving scenario to validate our method. In particular, we use data provided by a LiDAR sensor mounted on the roof of an autonomous vehicle. This data is part of the \textit{nuScenes} dataset \cite{CaeBanLanVorLioXuKriPanBalBei:20}.

\subsection{System Model and Implementation Details} \label{sec:exp_setup}

The nuScenes dataset consists of 1000 autonomous driving scenes and seven object classes. We use the official split of the dataset, where there are 700 scenes for training, 150 for validation, and 150 for testing. Each scene has a length of roughly 20 seconds and contains keyframes (frames with ground truth object annotations) sampled at 2Hz. Object detections extracted by the \textit{CenterPoint}\cite{YinZhoKra:21} detector are used as measurements, which are then preprocessed using \ac{nms} \cite{NeuVan:06}. Each measurement has a class index and the proposed \ac{mot} method is performed for each class individually.

We define the states of \acp{po} as $\V{x}_{k, n} \in \mathbb{R}^4$ which include their 2D position and 2D velocity. The measurements $\V{z}_{k, j} \in \mathbb{R}^5$ consist of the 2D position and velocity obtained as well as a score $0 \rmv<\rmv s_{k, j} \rmv\le\rmv 1$ from the object detector. 
The dynamics of objects are modeled by a constant-velocity model \cite{ShaKirLi:B02}. The \ac{roi} is given by $[x_e - 54, x_e + 54] \times [y_e - 54, y_e + 54]$, where $(x_e, y_e)$ is the 2D position of the autonomous vehicle. The prior \ac{pdf} of false alarms $f_{\text{fa}}(\cdot)$ and newly detected objects $f_{\text{n}}(\cdot)$ are uniform over the \ac{roi}. All other parameters used in the system model are estimated from the training data. The thresholds for object declaration was set to $T_{\text{dec}} \rmv=\rmv 0.5$ for legacy \acp{po} and to a class dependent value for new \acp{po}. The pruning threshold was set to $T_{\text{pru}} \rmv=\rmv 10^{-3}$.

The neural networks $g_{\text{e}}(\cdot)$, $g_{\text{n}}(\cdot)$, $g_{\text{a}}(\cdot)$, $g_{\text{motion}}(\cdot) \rmv\triangleq\rmv g_{\text{motion}, 1}(\cdot) \rmv=\rmv g_{\text{motion}, 2}(\cdot)$ are \acp{mlp} with a single hidden layer and leaky ReLU activation function. Furthermore, $g_{\text{r}}(\cdot)$ is a single-hidden-layer \ac{mlp} with sigmoid activation at the output layer. Finally, $g_{\text{shape}}(\cdot)  \rmv\triangleq\rmv g_{\text{shape}, 1}(\cdot)  \rmv= g_{\text{shape}, 2}(\cdot)$ consists of two convolutional layers followed by a single-hidden-layer \ac{mlp}. At each time frame, we use the output $\Set{Z}$ of \textit{VoxelNet} \cite{ZhuTuz:18} to extract shape features as discussed in Section \ref{sec:feaExtraction}. The used VoxelNet has been pre-trained by the CenterPoint method\cite{YinZhoKra:21}. Its parameters remain unchanged during the training of the proposed \ac{nebp} method. \ac{nebp} training is performed by employing the \textit{Adam optimizer} \cite{KinBa:14}. The number of \ac{gnn} iteration is $L = 3$. The batch size was set to $1$, the learning rate to $10^{-4}$, and the number of training epochs to 8. The tuning parameter $\epsilon$ in \eqref{eq:loss_beta} was set to\vspace{0mm} 0.1 and the threshold $T_{\text{dist}}$ for the pseudo ground truth extraction discussed in Section \ref{sec:Loss} was\vspace{.5mm} set to 2 meters.


\subsection{Performance Evaluation}

We use the widely used CLEAR metrics \cite{BerSti:08} that include the number of \ac{fp}, \ac{ids} and \ac{frag}. In addition, we also consider the \ac{amota} metric proposed in \cite{WenWanHelKit:20}. Note that the \ac{amota} is also the primary metric used for the nuScenes tracking challenge \cite{CaeBanLanVorLioXuKriPanBalBei:20}. 

Evaluation of the \ac{amota} requires a score for each estimated object. It was observed that a high \ac{amota} performance is obtained by calculating the estimated object score as a combination of existence probability and measurement score. In particular, for legacy \ac{po} $i$ the estimated object score is calculated\vspace{.5mm} as
\begin{equation}
 \underline{s}{}_i = \tilde{p}(\underline{r}{}_i = 1) + \sum_{j = 1}^{J} \ist \tilde{p}_{a_i}(j) \ist s_j, \nn
 \vspace{.5mm}
\end{equation}
where $\tilde{p}_{a_i}(j) \propto \phi_{a_i}(j) \kappa_{i}(j)$ are approximate marginal association probabilities \cite{MeyKroWilLauHlaBraWin:J18}. Finally, for new PO $j$ the estimated object score is given by $\overline{s}{}_j = \tilde{p}(\overline{r}{}_j = 1) + s_j $.
 
For a fair comparison, we use state-of-the-art reference methods that all rely on the CenterPoint detector \cite{YinZhoKra:21}. In particular, \ac{bp} refers to the traditional \ac{bp}-based \ac{mot} method \cite{MeyKroWilLauHlaBraWin:J18} that uses object detections provided by the CenterPoint detector as measurements. Furthermore, the tracking method proposed in \cite{YinZhoKra:21} uses a heuristic to create tracks and a greedy matching algorithm based on the Euclidean distance to associate CenterPoint object detections to tracks. Chiu et al. \cite{ChiLiAmbBoh:21} follows a similar strategy but makes use of a hybrid distance that combines the Mahalanobis distance and the so-called deep feature distance. Finally, the method introduced by Zaech et al. \cite{ZaeDaiLinDanVan:22} utilizes a network flow formulation and transforms the \ac{da} problem into a classification problem.

 In Table~\ref{tab:result}, it can be seen that the proposed \ac{nebp} approach outperforms all reference methods in terms of \ac{amota} performance. Furthermore, it can be observed, that \ac{bp} and \ac{nebp} achieve a much lower \ac{ids} and \ac{frag} metric compared to the reference methods. This is because both \ac{bp} and \ac{nebp} make use of a statistical model to determine the initialization and termination of tracks \cite{MeyKroWilLauHlaBraWin:J18} which is more robust compared to the heuristic track management performed by other reference methods. The improved \ac{amota} performance of \ac{nebp} over \ac{bp} comes at the cost of a slightly increased \ac{ids} and \ac{frag}.

TABLE \ref{tab:fp} shows the \ac{amota} performance as well as number of \ac{fp} for the bicycle and motorcycle class. To ensure a fair comparison, all the \ac{fp} values are evaluated for the same percentage of true positives referred to as ``recall''. In particular, for each class, the recall that leads to the largest multi-object tracking accuracy \cite{BerSti:08} for \ac{nebp} was used.

\begin{table}[!htbp] 
    \normalsize
    \centering
    \begin{tabular}{c|cc|cc}
    \hline
    \multirow{2}{*}{Method}         & \multicolumn{2}{c|}{bicycle}       & \multicolumn{2}{c}{motorcycle}     \\ \cline{2-5} 
                                    & AMOTA $\uparrow$ & FP $\downarrow$ & AMOTA $\uparrow$ & FP $\downarrow$ \\ \hline \hline
    CenterPoint \cite{YinZhoKra:21} & 0.458            & 390             & 0.615            & 792            \\
    BP                              & 0.505            & 168             & 0.725            & 349            \\
    NEBP (proposed)                 & \textbf{0.550}   & \textbf{120}    & \textbf{0.739}   & \textbf{208}   \\ \hline
    \end{tabular}
     \vspace{1mm}
    \caption{Evaluation results on nuScenes validation set in terms of \ac{amota} and \ac{fp} for the bicycle and motorcycle class.}
    \label{tab:fp}
\vspace{0mm}
\end{table}

For the considered two classes, \ac{nebp} yields the largest improvement in terms of \ac{amota} performance over \ac{bp}. Compared to \ac{bp}, the proposed \ac{nebp} method also has a reduced number of \ac{fp}. In conclusion, false alarm rejection and object shape association introduced by the proposed \ac{nebp} method can make effective use of features learned from raw sensor data and substantially improve \ac{mot} \vspace{1mm}performance.

\section{Final Remarks}
\vspace{0mm}

In this paper, we present a \ac{nebp} method for \ac{mot} that enhances probabilistic data association by features learned from raw sensor data. A \ac{gnn} is introduced that matches the topology of the factor graph for model-based data association. In addition to the preprocessed measurements employed by \ac{bp}, the \ac{gnn} also makes use of object features learned from raw sensor data. For false alarm rejection, the \ac{gnn} identifies which measurements are likely false alarms. For object shape association,  the \ac{gnn} computes improved association probabilities. The proposed method can improve the object declaration and state estimation performance of \ac{bp} while preserving its low computational complexity. Performance evaluation based on the nuScenes autonomous driving dataset demonstrated state-of-the-art object tracking performance.

\section*{Acknowledgement}
\vspace{-.5mm}

This work was supported by the National Science Foundation (NSF) under CAREER Award\vspace{0mm} No.~2146261. 

\renewcommand{\baselinestretch}{1.01}
\selectfont
\bibliographystyle{IEEEtran}
\bibliography{IEEEabrv,StringDefinitions,Books,Papers,ref}

\end{document}